%% file: iclr2027_conference.tex
\documentclass{article} 
\usepackage{iclr2027_conference,times}

\input{math_commands.tex}

\usepackage{hyperref}
\usepackage{helvet} 
\usepackage{courier}  
\usepackage{url}  
\usepackage{graphicx} 
\usepackage{amsmath}
\usepackage{amsfonts}
\usepackage{makecell}
\usepackage{algorithm}
\usepackage[noend]{algpseudocode}
\usepackage{subcaption}
\usepackage{booktabs}
\usepackage{wrapfig}

\title{Mixture of Decoders for Diverse Dialog Response Generation\thanks{ Preprints.} }

\author{Wenchao Du \\
Microsoft Corporation \\
Redmond, WA 98052 \\
\texttt{wenchaodu@microsoft.com} \\
}

\iclrfinalcopy 
\begin{document}

\maketitle

\begin{abstract}
Mixture modeling is a long established machine learning technique for learning large sets of multi-modal data. While it is known that sequence-to-sequence models for dialog response generation suffer from the problem of low diversity, we hypothesize that it is because sequence-to-sequence models tend to learn a degenerate uni-modal distribution of responses. We then propose to incorporate a mixture of decoders into sequence-to-sequence models and try to make each decoder learn specialized topics in order to improve the diversity of generated responses. Our model is developed under the framework of conditional variational autoencoder (CVAE). We evaluate our approach on an open domain chat corpus and show improvement over strong baselines in quantitative measures and human evaluation.
\end{abstract}

\section{Introduction}
Data-driven approaches to dialog response generation have become increasingly popular in the past few years \cite{ritter2011data}, and with the success of deep learning in language generation, especially machine translation \cite{sutskever2014sequence}, sequence-to-sequence models \cite{shang2015neural,vinyals2015neural} have started to dominate the field of dialog systems. While providing a high degree of automation and flexibility, deep learning models notoriously tend to generate dull and generic responses all the time \cite{li2016diversity}. Unlike machine translation, the complexity of response generation problem is very high in the way that each context (or source) may have multiple plausible responses, and those responses for the same context can be very different from each other in both lexical choices and syntactic structures. In other words, the output distribution for response generation is multi-modal, and its automation requires learning a one-to-many mapping between utterances \cite{liu2018towards}. Previous sequence-to-sequence learning approaches usually end up learning a degenerate output distribution whose mode is at the dull responses. This is widely known as ``modal collapse'' in machine learning community \cite{tolstikhin2017adagan}.

In order to learn such multi-modal distributions, we propose a general method combined with sequence-to-sequence learning, termed ``mixture of decoders (MoD)''. As suggested by the name, the model introduces a discrete latent variable, and each category is associated with a dedicated decoder for that mixture. Incorporating latent variables into encoder-decoder models to capture discourse-level information has been previously studied \cite{serban2017hierarchical,zhao2017learning}. We argue that mixture modelling has the following advantages. First, discrete latent variables are easier than continuous variables for interpretation, since mixtures can be directly mapped to proposed semantic or pragmatic categorizations, such as topics and dialog acts. Second, since the latent variables are discrete, it is made possible to incorporate prior knowledge of topics or dialog acts by means of semi-supervised learning, which is developed and discussed in this paper. Third, during the decoding process, a mixture of decoders can potentially provide a diverse pool of hypotheses, if the decoders are sufficiently specialized, and such diversity in the candidate pool leads to a higher chance in producing more diverse responses.

For learning and inference, we adopt the framework of conditional variational autoencoder \cite{sohn2015learning}. Although we only evaluate decoders based on recurrent neural networks (RNN), we conjecture that the mixture modelling approach can be applied to other types of decoders, such as convolutional neural networks \cite{zhang2017deconvolutional} and Transformer \cite{vaswani2017attention}. To our knowledge, we are the first to apply mixture modelling to dialog response generation problem with deep learning. From the perspective of machine learning, our model is an extension to M2 model of semi-supervised VAE \cite{kingma2014semi} by making the prior distribution conditioned on dialog contexts.

\section{Related Work}
Improving the diversity of dialog response generation and retrieval \cite{deb2019diversifying} has been an active research problem. \cite{li2016diversity} employed a decoding objective based on mutual information between contexts and responses; \cite{li2017learning} used reinforcement learning techniques for training the decoder to generate responses that maximize pre-defined rewards instead of perplexities; \cite{li2017adversarial,xu2017neural} adopted adversarial learning, in which a generator is trained to deceive a discriminator that tries to differentiate between generated responses and human responses. Beside changing training and decoding objectives, \cite{liu2018towards, lison2017not} considered reweighting data points by penalizing those with overly frequent responses or by emphasizing high-quality responses. \cite{du-black-2019-boosting} proposed a boosting-based approach for increasing the diversity by iteratively training the model with the data weighted differently at each iteration. Our approach aims to augment existing models and increase their capacity, and can be used together with aforementioned methods. Next, we survey in detail the related work on language model and generation from the following two aspects:

\setlength{\columnsep}{20pt}
\begin{wrapfigure}{r}{0.38\textwidth}
    \centering
    \includegraphics[width=\linewidth]{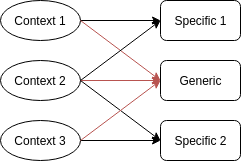}
    \caption{Sequence-to-sequence model learns a degenerate unimodal distribution of responses, as illustrated by the red lines.}
    \label{fig:degen}
\end{wrapfigure}

\subsection{Latent Variable Model} 
For dialog response generation, Serban et al. \cite{serban2017hierarchical} proposed to incorporate a continuous latent variable into their sequence-to-sequence model. The latent variable is conditioned on the dialog context, and the generation of responses depends on the same latent variable. Learning and inference of the latent variable involves a prior network conditioned on the context, and a posterior (or recognition) network conditioned on both the context and the response. Zhao et al. \cite{zhao2017learning} experimented with almost the same idea as Serban et al., in the hope that the continuous vector representations can be informative of discourse and pragmatics. They additionally developed a way to leverage linguistic knowledge by training the prior network and the recognition network with hand-crafted features. Du et al. \cite{du2018variational} advanced one step further in this line of ideas by learning separate latent variables for each word in the responses. Unlike these models with continuous latent variables, prior knowledge can be imparted onto a mixture model by training it discriminatively with labels. Discrete latent variables also allows better interpretability than continuous variables.

\subsection{Mixture Model for Language} 
There has been some recent work on combining mixture modelling and sequence-to-sequence learning for machine translation. He et al. \cite{he2018sequence} imposes a uniform prior on the mixtures. Shen et al. \cite{shen2019mixture} uses a multinomial distribution over mixtures that is agnostic to contexts, and update the parameters via expectation-maximization (EM). They also explored several variants on the learning procedure of their mixture model. In our work, the prior distribution of mixtures is conditioned on contexts, enabling the model to make sharper decisions of latent actions based on dialog history. 

Originally proposed for language modelling, mixture-of-softmax (MoS) \cite{yang2018breaking} aims to enhance the capacity of RNN decoders by learning a mixture of output word distributions on top of the bottleneck layer in the decoder. This constitutes a mixture model in which the weight of each mixture component is conditioned on the RNN hidden state at each time step. No posterior inference about the mixtures is made, and the parameters are learned through maximum likelihood estimation. We compare MoS with our approach in detail in the next section.

\begin{figure*}[th]
    \centering
    \includegraphics[scale=0.55]{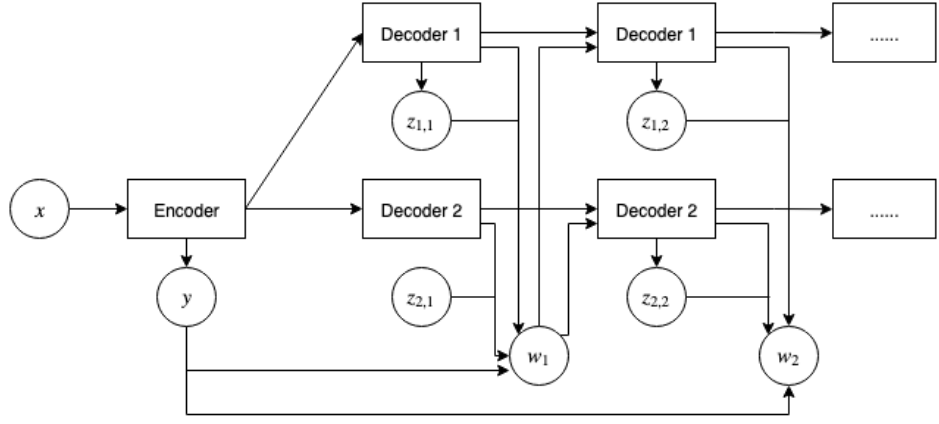}
    \caption{The graphical representation of our mixture of decoders model, where each decoder is a VAD. The figure shows the first two recurrent steps for two decoders. Circles are variables. Attention mechanism is left out for simplicity.}
    \label{fig:diagram}
\end{figure*}

\section{Methods}
\subsection{Formulation}
Given dialog context $x$, the standard sequence-to-sequence model learns to estimate the probability of response $\mathbf{w}$ through
\begin{equation*}
    p(\mathbf{w} \mid x) = \prod_i p_{\theta} (w_i \mid w_1 \dots w_{i-1}, x)
\end{equation*}
Previous work \cite{zhao2017learning} based on conditional variational autoencoder adds a continuous variable $z$ to connect $x$ and $\mathbf{w}$:
\begin{align*}
    p(\mathbf{w} \mid x) & = \int p_{\theta}(\mathbf{w}, z \mid x) dz \\
    & = \int p_{\theta}(\mathbf{w} \mid x, z) p_{\theta}(z \mid x) dz
\end{align*}
Variational Autoregressive Decoder (VAD) \cite{du2018variational} augment the idea by adding latent variables to each word in the response, and each latent variable depends not only on the context but also all previous words in the response:
\begin{align*}
    p(\mathbf{w} \mid x) & = \int p_{\theta}(\mathbf{w}, \mathbf{z} \mid x) dz \\
    & = \int \big ( \prod_i  p_{\theta}(w_i \mid z_i, w_1 \dots w_{i-1}, x) \cdot \\
    & p_{\theta} (z_i \mid w_1 \dots w_{i-1}, x) \big ) dz
\end{align*}
To add mixtures into the model, we introduce mixture variable $y$ and estimate the probability of responses through

\begin{align*}
    p(\mathbf{w} \mid x) & = \sum_y \int p_{\theta}(\mathbf{w}, \mathbf{z}, y \mid x) dz \\
    & = \sum_y p_{\theta}(y \mid x) \int p_{\theta}(\mathbf{w}, \mathbf{z} \mid x, y) dz
\end{align*}

Maximizing the objective above is intractable in general. We adopt mean-field variational inference and propose the variational posterior
\begin{equation*}
q_{\phi} (\mathbf{z}, y \mid \mathbf{w}, x) = q_{\phi} (y \mid \mathbf{w}, x) q_{\phi}(\mathbf{z} \mid \mathbf{w}, x, y)
\end{equation*}
and the parameters is learned through maximizing the evidence variational lower bound (ELBO):

\begin{align*}
\log & p_{\theta} (\mathbf{w} \mid x) \ge \\
& \mathbb{E}_q(y, \mathbf{z}) [\log p_{\theta} (\mathbf{w} \mid x, y, \mathbf{z}) + \log p_{\theta}(y \mid x) + \\
& \log p_{\theta} (\mathbf{z} \mid x, y, \mathbf{w}) - \log q_{\phi} (y, \mathbf{z} \mid x, \mathbf{w})]
\end{align*}

The prior distribution of $\mathbf{w}$ and $\mathbf{z}$ are parameterized as:
\begin{equation*}
p_{\theta} (y \mid x) \sim Cat(\pi_{\theta}(x))
\end{equation*}
\begin{equation*}
p_{\theta} (\mathbf{z} \mid x, y, \mathbf{w}) \sim \mathcal{N} (\mu_{\theta}(x, y, \mathbf{w}), diag(\sigma^2_{\theta}(x, y,  \mathbf{w}))
\end{equation*}
and the posterior distributions as:
\begin{equation*}
q_{\phi} (y \mid x, \mathbf{w}) \sim Cat(\pi_{\phi}(x, \mathbf{w}))
\end{equation*}
\begin{equation*}
q_{\phi} (\mathbf{z} \mid x, y, \mathbf{w}) \sim \mathcal{N} (\mu_{\phi}(x, y, \mathbf{w}), diag(\sigma^2_{\phi}(x, y, \mathbf{w}))
\end{equation*}

\subsection{Decoding}
To search for best hypotheses, we score each candidate with the marginal likelihood of its words. However, marginalizing the continuous variable $\mathbf{z}$ can be tricky when decoding. To keep the algorithm deterministic (rather than randomized) and fast, we choose to be greedy at each time step, using $\mathbf{z}$ with maximum likelihood, which equals the mean of the prior Gaussian distribution. To marginalize out $y$, we break down the search process into two stages:
\begin{itemize}
    \item \textbf{Nominate}. We perform normal beam search process for each mixture. The hypotheses are aggregated into one candidate pool.
    \item \textbf{Vote}. Each mixture scores all the candidates. Their scores are weighted according to the conditional prior of mixtures and added up for ranking candidates.
\end{itemize}

\begin{algorithm}
\caption{Pseudocode for decoding procedures.}
\begin{algorithmic}
\Procedure{Decode}{src}
\State $c \leftarrow []$
\For {each decoder $d$}
    \State $c \leftarrow c + BeamSearch(d, src)$
\EndFor
\For {$h$ in $c$}
    \State $h.score \leftarrow \sum_d P(h \mid src, d)P(d \mid src) $
\EndFor
\textbf{return} $h$ with highest score
\EndProcedure
\end{algorithmic}
\end{algorithm}

\subsection{Architectures}
We use bi-directional long short-term memory network \cite{schuster1997bidirectional,hochreiter1997long} to encode dialog history. Each mixture has its own LSTM for decoding. The last hidden states of the encoder, $\mathbf{h}_T^e$, are fed to a multi-layered perceptron (MLP) to obtain the initial hidden states of the decoder. Attention mechanism is used, and the previous hidden state of the decoder, $\mathbf{h}_{y, t-1}^{d}$, is used as the key to query encoder hidden states, and attention context, $\mathbf{c}_{y, t}$ is computed as the weighted sum of projected encoder hidden states.

\begin{equation*}
d_{y,t} = \langle \mathbf{W}_e \mathbf{h}_t^e,
\mathbf{W}_{y}^d \mathbf{h}_{y,T}^d \rangle
\qquad
e_{y,t} = \operatorname{softmax}(\mathbf{d}_y)
\qquad
\mathbf{c}_{y,t} = \sum_i e_{y,i}\mathbf{h}_t^e
\end{equation*}

The prior mixture probability is calculated with the last hidden states of the encoder:

\begin{equation*}
    \pi_{\theta}(y) = MLP_{\pi, \theta}(\mathbf{h}_T^e)
\end{equation*}

and the prior distribution of $\mathbf{z}$ at each time step $t$ are parameterized with the previous decoder hidden state and the current attention context:

\begin{equation*}
\begin{aligned}
        \mu_{\theta, y}(z_t) = MLP_{\mu, \theta, y}([\mathbf{h}_{t-1}^{d} ;  \mathbf{c}_{y, t}]) \\
        \log \sigma_{\theta, y}^2 (z_t) = MLP_{\sigma, \theta, y}([\mathbf{h}_{y, t-1}^{d} ;  \mathbf{c}_{y, t}])
\end{aligned}
\end{equation*}

We use an additional bi-directional LSTM encoder to encode responses into states $\mathbf{h}_t^p$ for posterior inference. The posterior distribution of mixtures is computed with the last hidden states of response encoder:

\begin{equation*}
    \pi_{\phi}(y) = MLP_{\pi, \phi}(\mathbf{h}_T^p)
\end{equation*}

and the posterior of $\mathbf{z}$ at each time step $t$ is computed with the next hidden states of response encoder:

\begin{equation*}
\begin{aligned}
    \mu_{\phi, y}(z_t) = MLP_{\mu, \phi, y}(\mathbf{h}_{t+1}^{p}) \\
    \log \sigma_{\phi, y}^2 (z_t) = MLP_{\sigma, \phi, y}(\mathbf{h}_{t+1}^{p})
\end{aligned}  
\end{equation*}

\subsection{Semi-Supervised Learning}
During experiments, we found degeneracy of latent mixtures to be an issue. This is similar to what is reported from \cite{shen2019mixture}. They reported two types of degeneracy:
\begin{enumerate}
    \item Only one decoder gets trained because ``rich get richer''. This means the decoder that learns fast will keep getting more weight. The other decoders end up being insufficiently trained.
    \item The discrete latent variable is ignored. The prior and posterior of mixtures often end up with uniform distribution, or a multinomial distribution that varies very little across different dialog contexts.
\end{enumerate}
In order to force the mixture model to make sharper decisions and be context-sensitive, we bootstrap labels to train the posterior network discriminatively. We do this by simply adding the log likelihood of labels to the training objective in Equation (2):
\begin{equation*}
    \mathcal{L}(\mathbf{w}, x, l) = \log q_{\phi} (y = l \mid \mathbf{w}, x)
\end{equation*}

The rest of the objective in (2) remains intact. Note that for labeled data, we still maximize the expected likelihood of words over all mixtures, instead of the likelihood from the labeled mixture. This is different from \cite{kingma2014semi}. We still allow each mixture to learn the data from other mixtures, but with lower weights. This is because we found through experiments that if trained with only a small subset of the data, the decoders of each mixture may generate responses with more specialized words, but in a highly monotonous way.

To bootstrap labels, we group utterances based on their lexical choices, in the hope that the resulting mixtures would reflect different topics. More specifically, we embed each utterance using the weighted sum of the Glove embeddings \cite{pennington2014glove} of words in the utterance, normalized by subtracting the largest singular vector of all embeddings, as described in \cite{arora2016simple}. Then we perform k-means clustering on these embeddings, with the same number of clusters as mixtures in the response generation model. In addition, we assign ``dull'' responses to one specific mixture. Responses are considered ``dull'' if it contains n-grams ($n \ge 4$) that are highly frequent in those responses generated by sequence-to-sequence baseline. The mixture is specially handled in training. It still has access to responses from other mixtures, but the other mixtures do not learn resposes that are labeled as ``dull''. 

\begin{table}[t!]
\small
\begin{center}
\begin{tabular}{|l|l|l|}
\hline  & \bf MoD & \bf MoS \\ \hline
\bf Output &  &  \checkmark \\
\bf Bottleneck & \checkmark &  \\
\bf Attention & \checkmark &  \\
\bf RNN Decoder & \checkmark  & \\

\hline
\end{tabular}
\end{center}
\caption{\label{one} Parameter sharing in different layers in mixture of decoders (MoD) and mixture of softmax (MoS). Check mark indicates each mixture uses its own parameters. }
\end{table}

\begin{table}[t!]
\small
\begin{center}
\begin{tabular}{|l|l|l|}
\hline  & \bf MoD & \bf MoS \\ \hline
\bf Order & $O(KEH)$ &  $O(KEV)$ \\
\bf K = 3 & $< 9M$ & $> 18M$ \\
\bf K = 5 & $< 11M$ & $ > 30M$ \\

\hline
\end{tabular}
\end{center}
\caption{\label{two} Comparison of the order and numeric values of the number of parameters used by MoD and MoS . }
\end{table}

\begin{table*}[h!t]
\small
\centering
\setlength{\tabcolsep}{3pt}
\begin{tabular}{|c|c|c|c|c|c|c|c|}
\hline
& Human & Seq2Seq & VAD & MoS & MoD & \makecell{MoD \\ (without z)} & \makecell{MoD \\ (Supervised) } \\
\hline
Length & 10.35 & 8.93 & 7.88 & \textbf{10.53} & 12.34 &  9.24 & 14.01 \\
BLEU & N/A & 0.135 & 0.126 & 0.159 & \textbf{0.169} & 0.152 & 0.157 \\
ROUGE-L & N/A & 0.155 & 0.153 & 0.164 & \textbf{0.170} & 0.158 & 0.160 \\
\hline
Cos. Sim. to targets & 1.0000 & 0.0481 & 0.0603 & 0.0611 & 0.0699 & \textbf{0.0820} & 0.0715 \\
Cos. Sim. to persona & 0.1246 & 0.0614 & 0.0480 & 0.0491 & 0.0668 & 0.0515 & \textbf{0.0810} \\
\hline
Distinct uni-gram & 6327 & 353 & 566 & 424 & 617 & 590 & \textbf{661} \\
Distinct bi-gram & 44376 & 931 & 1559 & 1264 & 2106 & 1866 & \textbf{2413} \\
Distinct tri-gram & 77884 & 1499 & 2437 & 2159 & 3744 & 3224 & \textbf{4415} \\
\hline
\end{tabular}
\caption{Evaluation results on Persona dataset.}
\label{tab:sixth}
\end{table*}

\subsection{Model Comparison}
We now discuss the distinction between mixture of decoders and mixture of softmax in detail. The focus is on the efficient use of parameters. In mixture of decoders, the last hidden state of decoder, the attention context, and the continuous latent variable are used for word prediction. The concatenations of aforementioned representations are first passed through a bottleneck layer (for the sake of reducing the number of parameters):
\begin{equation*}
    \mathbf{b}_{y, t} = tanh(\mathbf{W}_b[\mathbf{h}_{d, y}^{t} ;  \mathbf{c}_{y, t}; \mathbf{z}_{y, t}])
\end{equation*}
The size of bottleneck is usually chosen to be the size of word embeddings. Then the output is used for calculating the distribution over words
\begin{equation*}
    p(w_t) = softmax(\mathbf{E} \mathbf{b}_{y, t})
\end{equation*}
Note that the weight matrix in the output layer is shared across mixtures, for reducing the number of parameters to learn. On the other hand, mixture of softmax first predicts mixture probabilities from the output of the bottleneck layer
\begin{equation*}
    \mathbf{\pi}_{y, t} = softmax(\mathbf{W}_{\pi}\mathbf{b}_t)
\end{equation*}
Then the probability of words are marginalized over mixtures
\begin{equation*}
    p(w_t) = \sum_y \pi_{y, t} softmax(\mathbf{E}_y \mathbf{b}_t)
\end{equation*}
with each mixture using its own weights in the final output layer. The difference of how mixture of decoders and mixture of softmax share weights is summarized in Table 1.

Next, we compare the number of parameters used by mixture of decoders and mixture of softmax. Let $V$ be the size of vocabulary, $E$ be the size of word embeddings, $K$ be the number of mixtures, and $H$ be the size of any other hidden variables including attention contexts and decoder states. We calculate the order of number of parameters used by both models, and the actual values when the number of mixtures is at 3 and 5. The results are summarized in Table 2. It shows the number of parameters used by mixture of decoders is much smaller; the number of parameters used by mixture of softmax grows very fast with the number of mixtures.

Now we can see how mixture of decoders is advantageous over mixture of softmax. By utilizing distinct set of parameters for each mixture across multiple layers, mixture of decoders learns a rich set of latent representations while keeping the number of parameters small.

\subsection{Implementation and Training Details}
All models are implemented using PyTorch. The hidden size of all LSTM encoders and decoders are 512. The size of embeddings of words is 300. We trained our models using Adam optimization algorithm \cite{kingma2014adam} with learning rate $0.001$. Dropout \cite{srivastava2014dropout} of rate $0.25$ is used wherever applicable. Batch size is adjusted according to the number of mixtures. We trained our model on a single GPU (GTX 1080 Ti). We observed that the training speeds of mixture of decoders and mixture of softmax are usually 3 to 5 times slower than baseline models without mixtures.

\begin{table*}[t]
\centering
\small
\begin{tabular}{lp{0.82\textwidth}}
\toprule
\textbf{Context} &
Sorry to hear that. Making crafts takes my stress away. I enjoy it. \\
\textbf{Persona} &
I love animals and have two dogs. \\
\textbf{Baseline} &
What do you do for a living? \\
\textbf{MoD} &
Do you like dogs? I love them. \\
\midrule
\textbf{Context} &
I grew up with Morgans. I love horses. \\
\textbf{Persona} &
I also enjoy cooking meals with food from our garden. \\
\textbf{Baseline} &
I do not have any pets. Do you have any pets? \\
\textbf{MoD} &
Do you like cooking? I love to bake. \\
\midrule
\textbf{Context} &
Great! How are you? \\
\textbf{Persona} &
I am married and have two kids of my own. \\
\textbf{Baseline} &
I am good, just got back from the gym. \\
\textbf{MoD} &
I am great. Just hanging out with my wife and two kids. \\
\bottomrule
\end{tabular}
\caption{Positive examples of generated responses from the MoD model with better reference to persona.}
\label{tab:third}
\end{table*}

\section{Experiments and Analysis}
\subsection{Data}
We evaluate our model for dialog response generation on Persona dataset \cite{zhang2018personalizing}, an open-domain dialog corpus. Each person is give a list of persona descriptions in simple sentences, and they are required to converse according to the given persona. We use the last utterance for each response as source, with persona descriptions prepended. During pre-processing, we truncate all trailing punctuations. The corpus has a vocabulary around 20,000 words.

\subsection{Quantitative Analysis}
We use the following sets of metrics for automatic evaluation of the quality of generated responses.

\subsubsection{Count Statistics} 
We show the average length of generated responses. To measure lexical diversity, we count the number of distinct n-grams in all the generated responses.

\subsubsection{Word Overlap Based Metrics} We use BLEU \cite{papineni2002bleu} and ROUGE-L \cite{lin2004rouge} scores as automatic evaluation metrics. While the reliability of such metrics has been criticized \cite{liu2016not}, there is also evidence supporting the correlation of these metrics with human judgment for task-oriented domains \cite{sharma2017relevance}.

\subsubsection{Word Embedding Based Metrics} We evaluate the semantic similarity between generated responses and human responses and associated persona descriptions by the cosine distance of their sentence embeddings. We use the word averaging approach by \cite{arora2016simple} to embed the responses, which has been demonstrated to be very good at capturing lexical level semantic similarity.

The main results are listed in Table 3. The number of mixtures for MoS and MoD is 3. We also consider the baseline in which we remove the continuous latent variable from MoD. BLEU scores are correlated with lengths due to its brevity penalty. We found that MoS generate responses that have closest lengths to human responses on average. VAD tend to generate short responses. MoDs tend to generate longer responses. On manual inspection, we found that a fraction of those long responses generated by MoDs are self-repeating. As for metrics based on word overlapping, MoD beat baseline models. Adding continuous latent variables to MoD helps improve these metrics.

When it comes to semantic similarity based on embeddings, unsupervised MoD with discrete latent variables only performed the best on matching ground truth responses, while supervised MoD achieved the best result on matching persona descriptions. Full MoD also performed well comparing against baseline models.

As for lexical diversity. we can see that by adding either discrete or continuous latent variables into encoder-decoder models, better diversity is achieved, since VAD, MoS, and MoD all outperformed sequence-to-sequence baseline. Adding continuous latent variables helps improve lexical diversity, while adding supervision into MoD can make further improvement. 
\begin{figure*}
\centering
  \begin{subfigure}[b]{0.31\textwidth}
    \includegraphics[width=\textwidth]{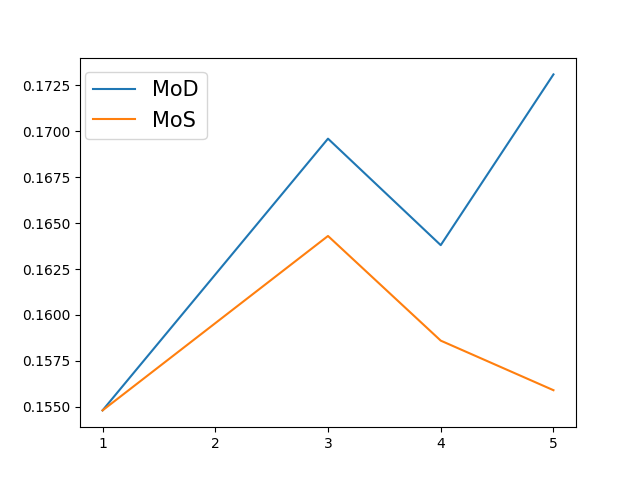}
    \caption{ROUGE}
    \label{fig:1}
  \end{subfigure}
  \begin{subfigure}[b]{0.31\textwidth}
    \includegraphics[width=\textwidth]{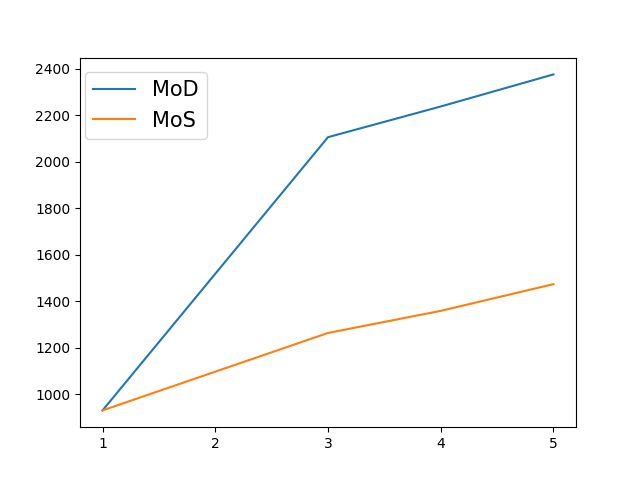}
    \caption{Number of bi-grams}
    \label{fig:2}
  \end{subfigure}
\begin{subfigure}[b]{0.31\textwidth}
    \includegraphics[width=\textwidth]{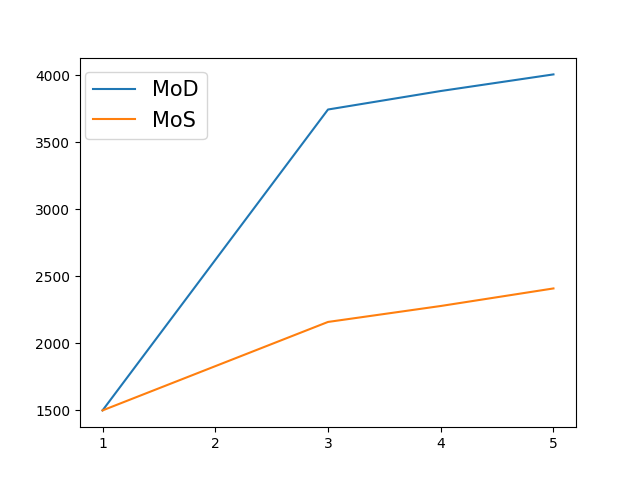}
    \caption{Number of tri-grams}
    \label{fig:2}
  \end{subfigure}
\caption{Quantitative results by varying the number of mixtures. X-axis is for number of mixtures and y-axis for metrics. The number of mixtures are experimented with K = 1, 3, 4, 5. }
\label{fig:test}
\end{figure*}

We also experimented with the effects of number of mixtures on MoS and MoD. The maximum number of mixtures is limited to 5 due to GPU memory constraint. Results are summarized in Figure 3. The general trend is that when the number of mixtures is small, increasing the number may improve lexical diversity, but not always help matching words in references, as shown in ROUGE-L. Note the trend that ROUGE-L starts to decline for MoS at $K =3$.

\subsection{Qualitative Analysis}

\begin{table*}[ht]
\small
\begin{center}
\begin{tabular}{|l|l|l|l|}
\hline \textbf{Model} & \textbf{Win} & \textbf{Loss} & \textbf{Tie} \\ \hline
Ours vs VAD & $47.0 \pm 12.1 \%$ & $25.0 \pm 10.3 \%$ & $28.0 \pm 19.8 \%$ \\ \hline
Ours vs MoS & $42.0 \pm 12.2 \%$ & $30.0 \pm 6.6 \%$ & $28.0 \pm 11.0 \%$ \\ \hline
\end{tabular}
\end{center}
\caption{\label{human} Human evaluation results. }
\end{table*}

\begin{table*}[h!t]
\small
\centering
\setlength{\tabcolsep}{3pt}
\begin{tabular}{|c|c|c|c|}
\hline
 & \bf Example 1 & \bf Example 2 & \bf Example 3 \\
\hline
\bf Mix. 1 &  What kind of music do you like & \makecell{Do you like movies ? \\ I love horror movies} & Do you like to watch TV \\
\hline
\bf Mix. 2 & What city are you from & \makecell{Yes I do . \\ I am a full time student} & What are you majoring in \\
\hline
\bf Mix. 3 & Hi how are you today & I am great . How are you & That sounds like a lot of fun \\
\hline
\bf Mix. 4 & \makecell{I am a huge fan of red hot chili \\ peppers . They are my favorite} & \makecell{I am a huge fan of \\ French fries . You} & \makecell{Do you like to cook ? \\ I love it} \\
\hline
\bf Mix. 5 & What do you do for a living & \makecell{What do you like to do \\ in your spare time} & \makecell{I do not have a lot of \\ free time . What about you} \\
\hline
\end{tabular}
\caption{Examples of generated responses from each mixture in supervised MoD.}
\label{tab:third}
\end{table*}

While the mixture model can improve word diversity, we need to verify the diversified responses are still as relevant. We ask 4 annotators to evaluate a randomly sampled subset of 100 examples. We compare our model to two baseline models: VAD and MoS with same number of mixtures. Each context is paired with two responses -- one from the baseline model and one from our model. The annotators are asked to choose the best response, based on (1) appropriateness, and (2) relevance to given persona descriptions. Ties are allowed. The results are shown in Table 5. On average, about 30 percent of the time the annotators showed no preferences, and our mixture of decoders model are preferred over VAD and MoS for more than 40 percent of the time, and lost less than 30 percent of the trials. Note that all individual tests show annotators preferred MoD over baseline models, except for one case, where the annotator chose MoS over MoD slightly more often. We also show some of the examples on how MoD did well in Table 4.

To better understand what each mixture is learning about semantic topics, in Table 4, we select some examples from the most frequently generated responses for each mixtures. It can be observed that Mixture 1 tend to focus on hobbies, especially entertainment. Mixture 2 covers education. Mixture 3 contains conversation starters. Mixture 4 specifically concentrates on food. Mixture 5 tend to give general inquiries of occupation and leisure.

To evaluate how diverse are the responses generated by different mixtures, we follow \cite{shen2019mixture} and use pairwise BLEU to score the hypotheses set. The lower the pairwise BLEU scores, the more diverse the hypotheses. Additionally, we count the average number of distinct n-grams of hypotheses. We compare supervised MoD of 5 mixtures against the other baseline models. The top 5 hypotheses from the baseline are used, while the top hypothesis from each mixture in MoD are selected. The results are listed in Table 7. The sequence-to-sequence baseline has the highest pairwise BLEU. VAD, MoS, unsupervised MoD have slightly lower pairwise BLEU and slightly better lexical diversity in hypotheses sets. With supervision, pairwise BLEU score significantly reduced for MoD, indicating each mixture decoder became more specialized. 

\begin{table*}[h!t]
\small
\centering
\begin{tabular}{|c|c|c|c|c|c|}
\hline
& Seq2Seq & VAD & MoS & MoD & \makecell{MoD \\ (Supervised) } \\
\hline
BLEU & 0.58 & 0.49 & 0.47 & 0.46 & \textbf{0.22} \\
\hline
Distinct uni-gram & 22.8 & 22.3 & 23.4 & 23.8 & \textbf{24.5} \\
Distinct bi-gram & 59.7 & 58.7 & 70.1 & 75.4 & \textbf{86.9} \\
Distinct tri-gram & 76.4 & 74.0 & 98.0 & 106.7 & \textbf{122.9} \\
\hline
\end{tabular}
\caption{Diversity evaluation of hypotheses sets. }
\label{tab:sixth}
\end{table*}

\section{Conclusion}
In this paper we investigate the idea of combining mixture modeling with sequence-to-sequence model to improve the diversity issue in automated dialog response generation. Our model is developed under the framework of autoencoding variational Bayes. A discrete latent variable is introduced to capture latent dialog actions or topics. Such latent variable is easy to interpret. As illustrated in our qualitative analysis, each mixture specializes in certain topics to some extent. We also demonstrated the possibility of leveraging prior knowledge for learning latent topics through semi-supervised learning. By examining the hypotheses set, we confirm that the source of the superior diversity of mixture of decoders model is indeed the specialization of decoders. Our model performed reasonably well, comparing against other latent variable models and mixture models, in terms of automatic metrics. Moreover, our claim is backed by human evaluation through A/B test, and the responses generated by our model is often preferred over baseline models. As a future direction, we believe that a more sophisticated way (with more fine-grained linguistic features) to supervised the mixture of decoders model can further improve its performance.

\bibliography{reference}
\bibliographystyle{iclr2027_conference}

\end{document}

%% file: math_commands.tex
\usepackage{amsmath,amsfonts,bm}

\def\eqref#1{equation~\ref{#1}}

\def\1{\bm{1}}

\DeclareMathAlphabet{\mathsfit}{\encodingdefault}{\sfdefault}{m}{sl}
\SetMathAlphabet{\mathsfit}{bold}{\encodingdefault}{\sfdefault}{bx}{n}

